\documentclass[10pt,twocolumn,letterpaper]{article}
\pdfoutput=1
\usepackage{cvpr}
\usepackage{times}
\usepackage{graphicx}
\usepackage{amsmath}
\usepackage{amssymb}
\usepackage[normalem]{ulem}

\usepackage{authblk}
\usepackage{multirow}

\graphicspath{ {./images/} }
\cvprfinalcopy
\def\cvprPaperID{1031}

\ifcvprfinal\fi
\begin{document}
%
%
%
\title{Monocular Real-time Hand Shape and Motion Capture using Multi-modal Data}
%
%
\renewcommand\Authands{ }
\renewcommand\Authand{ }
\renewcommand\Authsep{ }
\renewcommand\Authfont{}
\renewcommand\Affilfont{\small}

\author[1]{Yuxiao Zhou}
\author[2,3]{Marc Habermann}
\author[2,3]{Weipeng Xu}
\author[2,3]{Ikhsanul Habibie}
\author[2,3]{Christian Theobalt}
\author[1]{Feng Xu\thanks{
This work was supported by the National Key R\&D Program of China 2018YFA0704000, the NSFC (No.61822111, 61727808, 61671268), the Beijing Natural Science Foundation (JQ19015, L182052), and the ERC Consolidator Grant 4DRepLy (770784).
Feng Xu is the corresponding author.}}

\affil[ ]{
  \textsuperscript{1}BNRist and School of Software, Tsinghua University,
  \textsuperscript{2}Max Planck Institute for Informatics,
  \textsuperscript{3}Saarland Informatics Campus
}
%
%
\maketitle
%
%
\begin{abstract}
We present a novel method for monocular hand shape and pose estimation at unprecedented runtime performance of 100fps and at state-of-the-art accuracy.
This is enabled by a new learning based architecture designed such that it can make use of all the sources of available hand training data: image data with either 2D or 3D annotations, as well as stand-alone 3D animations without corresponding image data.
It features a 3D hand joint detection module and an inverse kinematics module which regresses not only 3D joint positions but also maps them to joint rotations in a single feed-forward pass.
This output makes the method more directly usable for applications in computer vision and graphics compared to only regressing 3D joint positions.
We demonstrate that our architectural design leads to a significant quantitative and qualitative improvement over the state of the art on several challenging benchmarks.
Our model is publicly available for future research.\footnote{https://github.com/CalciferZh/minimal-hand}
\end{abstract}
%
%
%
\section{Introduction}
\label{sec:intro}
%
%
\begin{figure}[t]
    \centering
    \includegraphics[width=\linewidth]{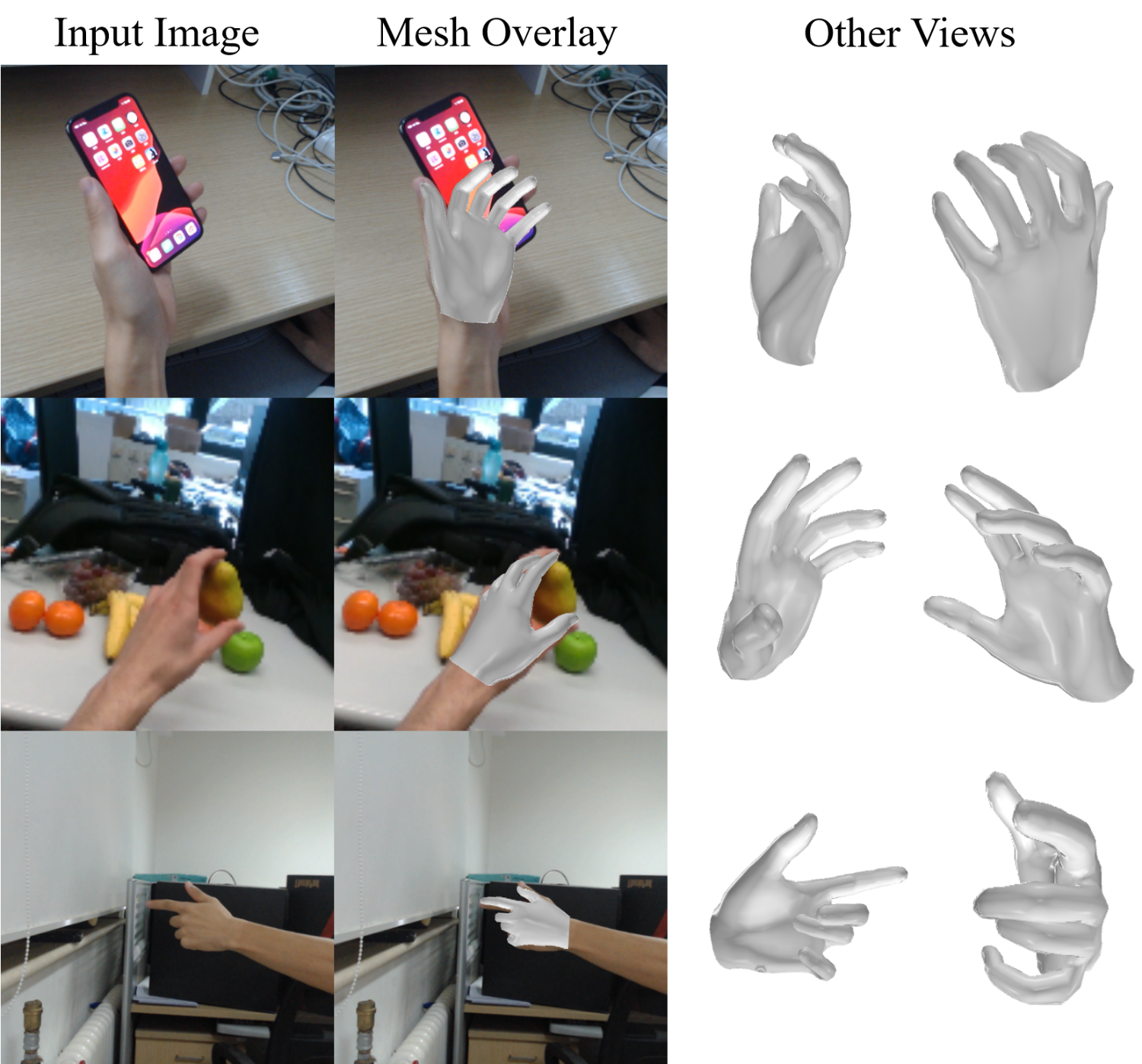}
	\caption{
		We present a novel hand motion capture approach that estimates 3D hand joint locations and rotations at real time from a single RGB image.
		Hand mesh models can then be animated with the predicted joint rotations.
		Our system is robust to challenging scenarios such as object occlusions, self occlusions, and unconstrained scale.
	}
	\label{fig:teaser}
	\vspace{-4mm}
\end{figure}
%
%
%
\par
Hands are the most relevant tools for humans to interact with the real world.
Therefore, capturing hand motion is of outstanding importance for a variety of applications in AR/VR, human computer interaction, and many more.
Ideally, such a capture system should run at real time to provide direct feedback to the user, it should only leverage a single RGB camera to reduce cost and power consumption, and it should predict joint angles as they are more directly usable for most common applications in computer graphics, AR, and VR.
3D hand motion capture is very challenging, especially from a single RGB image, due to the inherent depth ambiguity of the monocular setting, self occlusions, complex and fast movements of the hand, and uniform skin appearance.
The existing state-of-the-art methods resort to deep learning and have achieved significant improvement in recent years~\cite{Boukhayma2019CVPR,Ge2019CVPR,Zhang2019ICCV,Iqbal2018ECCV,Cai2018ECCV}.
However, we observe two main problems with those methods.
%
%
\par 
%
%
First, none of the existing methods makes use of all publicly available training data modalities, even though the annotated hand data is severely limited due to the difficulty of collecting real human hand images with 3D annotations.
Specifically, to obtain 3D annotations, a particular capture setup is required, e.g., leveraging stereo cameras \cite{Zhang2017ICIP} or a depth camera \cite{Sridhar2016ECCV,Tompson2014ToG,Tang2014CVPR,Yuan2017CVPR}, which prevents collecting diverse data at large scale.
An alternative is synthetic datasets \cite{Mueller2018CVPR,Zimmermann2017ICCV}.
However, models trained on synthetic images do not generalize well to real images due to the domain gap \cite{Mueller2018CVPR}.
In contrast, 2D annotated internet images with larger variation \cite{Simon2017CVPR} are easier to obtain.
However, it is nearly impossible to annotate them with the 3D ground truth.
We notice that, there is another valuable data modality neglected by all previous works - hand motion capture (MoCap) data.
These datasets usually have a large variation in hand poses, but lack the paired images, since they are typically collected using data gloves~\cite{Glauser2019ToG} or 3D scanners~\cite{Romero2017ToG}.
Therefore, the previous methods cannot use them to learn the mapping from images to hand poses.
%
%
\par 
%
%
Second, most previous methods focus on predicting 3D joint positions \cite{Yang2019ICCV,Spurr2018CVPR,Iqbal2018ECCV,Cai2018ECCV,Zimmermann2017ICCV}.
Although useful for some applications, this positional representation is not sufficient to animate hand mesh models in computer graphics, where joint rotations are typically required.
Some works \cite{Mueller2018CVPR,Panteleris2018WACV,Tompson2014ToG} overcome this issue by fitting a kinematic hand model to the sparse predictions as a separate step.
This not only requires hand-crafted energy functions, but expensive iterative optimization also suffers from erroneous local convergence.
Other works \cite{Zhang2019ICCV,Baek2019CVPR,Boukhayma2019CVPR} directly regress joint angles from the RGB image.
All of them are trained in a weakly-supervised manner (using differentiable kinematics functions and the 3D/2D positional loss) due to the lack of training images paired with joint rotation annotations.
Therefore, the anatomic correctness of the poses cannot be guaranteed.
%
%
\par 
%
%
To this end, we propose a novel real-time monocular hand motion capture approach that not only estimates 2D and 3D joint locations, but also maps them directly to joint rotations.
Our method is rigorously designed for the utilization of all aforementioned data modalities, including synthetic and real image datasets with either 2D and/or 3D annotations as well as non-image MoCap data, to maximize accuracy and stability.
Specifically, our architecture comprises two modules, \textit{DetNet} and the \textit{IKNet}, which predict 2D/3D joint locations and joint rotations, respectively.
%
%
The proposed DetNet is a multi-task neural network for 3D hand joint detection that can inherently leverage fully and weakly annotated images at the same time by explicitly formulating 2D joint detection as an auxiliary task.
In this multi-task training, the model learns how to extract important features from real images leveraging 2D supervision, while predicting 3D joint locations can be purely learned from synthetic data.
%
%
The 3D shape of the hand can then be estimated by fitting a parametric hand model \cite{Romero2017ToG} to the predicted joint locations.
%
%
To obtain the joint rotation predictions, we present the novel data-driven end-to-end IKNet that tackles the inverse kinematics (IK) problem by taking the 3D joint predictions of DetNet as input and regressing the joint rotations.
Our IKNet predicts the kinematic parameters in a single feed-forward pass at high speed and it avoids complicated and expensive model fitting.
During training, we can incorporate MoCap data that provides direct rotational supervision, as well as 3D joint position data that provides weak positional supervision, to learn the pose priors and correct the errors in 3D joint predictions.
In summary, our contributions are:
\begin{itemize}
	\item{A new learning based approach for monocular hand shape and motion capture, which enables the joint usage of 2D and 3D annotated image data as well as stand-alone motion capture data.}
	\item{An inverse kinematics network that maps 3D joint predictions to the more fundamental representation of joint angles in a single feed-forward pass and that allows joint training with both positional and rotational supervision.}
\end{itemize}
Our method outperforms state-of-the-art methods, both quantitatively and qualitatively on challenging benchmarks, and shows unseen runtime performance.
\section{Related Work}
In the following, we discuss the methods that use a single camera to estimate 3D hand pose, which are closely related to our work.
%
%
\par \noindent \textbf{Depth based Methods.}
Many works proposed to estimate hand pose from depth images due to the wide spread of commodity depth cameras.
Early depth based works \cite{Oikonomidis2011BMVC,Melax2013GI,Schroder2014ICRA,Fleishman2015CVPRW,Tagliasacchi2015CGF,Tkach2016ToG} estimate hand pose by fitting a generative model onto a depth image.
Some works \cite{Sridhar2015CVPR,Sharp2015CHI,Taylor2017ToG,Sridhar2016ECCV,Tzionas2016IJCV} additionally leveraged discriminative predictions for initialization and regularization.
Recently, deep learning methods have been applied to this area.
As a pioneer work, Tompson et al.~\cite{Tompson2014ToG} proposed to used CNN in combination with randomized decision forests and inverse kinematics to estimate hand pose from a single depth image at real time.
Follow-up works achieved better performance by utilizing priors and context \cite{Markus2015CVWW}, high-level knowledge \cite{Tang2019TPAMI}, a feedback loop \cite{Markus2015ICCV,Markus2019TPAMI}, or intermediate dense guidance map supervision \cite{Wu2018ECCV}.
\cite{Zhou2018ECCV,Du2019CVPR} proposed to use several branches to predict the pose of each part, e.g. palm and fingers, and exploit cross-branch information.
Joint estimation of hand shape and pose was also proposed \cite{Malik20183DV}.
Wan et al.~\cite{Wan2019CVPR} exploited unlabeled depth maps for self-supervised finetuning, while Mueller et al. \cite{Mueller2017ICCV} constructed a photorealistic dataset for better robustness.
Some works leveraged other representations, such as point clouds \cite{Ge2018CVPR, Ge2018ECCV, Chen2019ICCV, Li2019CVPR} and 3D voxels \cite{Huang2018BMVC,Moon2018CVPR,Ge2019TPAMI}, which can be retrieved from depth images.
Although these works achieve appealing results, they suffer from the inherent drawbacks of depth sensors, which do not work under bright sunlight, have a high power consumption and people have to be close to the sensor.
%
%
\par \noindent \textbf{Monocular RGB Methods.}
To this end, people recently started to research 3D hand pose estimation from monocular RGB images, which is even more challenging than the depth based setting due to the depth ambiguity.
Zimmermann and Brox~\cite{Zimmermann2017ICCV} trained a CNN based model that estimates 3D joint coordinates directly from an RGB image.
Iqbal et al.~\cite{Iqbal2018ECCV} used a 2.5D heat map formulation, that encodes 2D joint locations together with depth information, leading to a large boost in accuracy.
For better generalization, many works \cite{Cai2018ECCV,Spurr2018CVPR,Yang2019ICCV} utilized depth image datasets to enlarge the diversity seen during training.
Mueller et al.~\cite{Mueller2018CVPR} proposed a large scale rendered dataset post-processed by a CycleGAN\cite{Zhu2017ICCV} to bridge the domain gap.
However, they only focused on joint position estimation but refrained from joint rotation recovery, which is much better for hand mesh animation.
To estimate joint rotations, \cite{Xiang2019CVPR,Panteleris2018WACV} fitted a generic hand model to the predictions via an iterative optimization based approach, which is not time-efficient and requires hand-crafted energy functionals.
\cite{Zhang2019ICCV,Baek2019CVPR} proposed to regress the parameters of a deformable hand mesh model from the input image in an end-to-end manner.
Nonetheless, the estimated rotations can only be weakly supervised, resulting in inferior accuracy.
Ge et al.~\cite{Ge2019CVPR} directly regressed a hand mesh using a GraphCNN \cite{Defferrard2016NIPS}, but a special dataset with ground truth hand meshes is required, which is hard to construct.
Their model-free method is also less robust to challenging scenes.
In contrast, by fully exploiting existing datasets from different modalities, including image data and non-image MoCap data, our approach obtains favorable accuracy and robustness.

\section{Method}
%
%
\begin{figure*}[t]
	\centering
	\includegraphics[width=\textwidth]{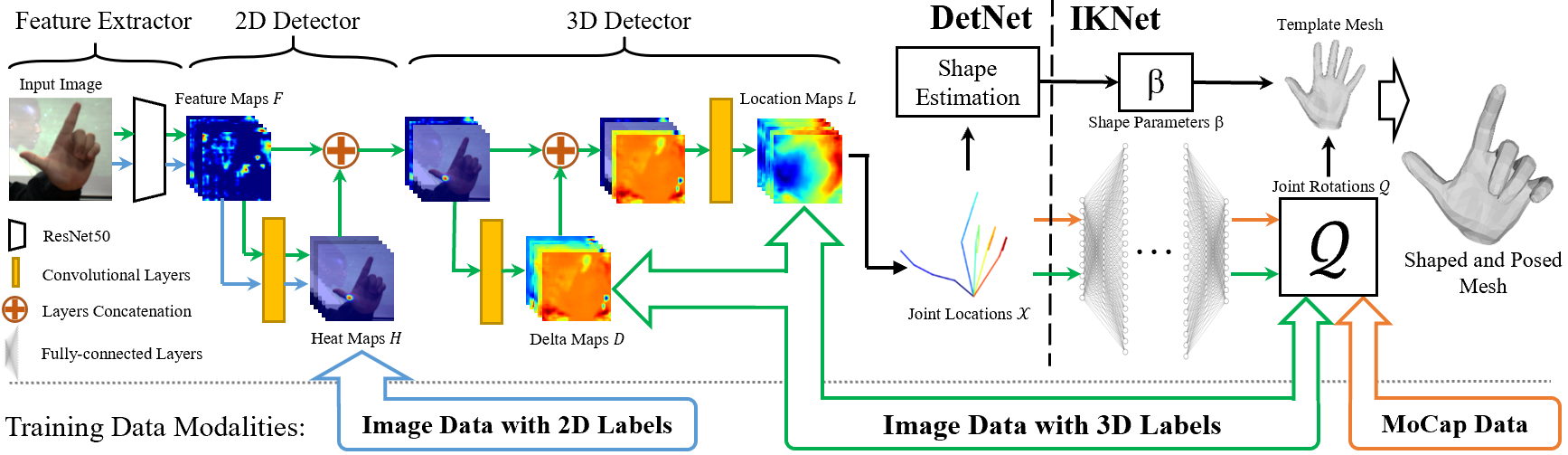}
	\caption{
    Overview of our architecture.
    It comprises two modules:
    first, our DetNet predicts the 2D and 3D joint positions from a single RGB image.
    Second, our IKNet takes the 3D joint predictions of DetNet and maps them to joint angles.
	}
	\label{fig:overview}
	\vspace{-2mm}
\end{figure*}
%
%
As shown in Fig.~\ref{fig:overview}, our method includes two main modules.
First, the joint detection network, DetNet (Sec.~\ref{sec:method:detnet}), predicts 2D and 3D hand joint positions from a single RGB image under a multi-task scheme.
Then, we can retrieve the shape of the hand by fitting a hand model to the 3D joint predictions (Sec.~\ref{sec:handmodel_shape}).
Second, the inverse kinematics network, IKNet (Sec.~\ref{sec:method:iknet}), takes the 3D joint predictions and converts them into a joint rotation representation in an end-to-end manner.
\subsection{Hand Joint Detection Network DetNet}
\label{sec:method:detnet}
The DetNet takes the single RGB image and outputs root-relative and scale-normalized 3D hand joint predictions as well as 2D joint predictions in image space.
The architecture of DetNet comprises 3 components: a feature extractor, a 2D detector, and a 3D detector.
%
%
\par \noindent \textbf{Feature Extractor.}
We use the backbone of the ResNet50 architecture \cite{He2016CVPR} as our feature extractor where the weights are initialized with the Xavier initialization \cite{Glorot2010AIS}.
It takes images at a resolution of $128 \times 128$ and outputs a feature volume $F$ of size $32 \times 32 \times 256$.
%
%
\par\noindent \textbf{2D Detector.}
The 2D detector is a compact 2-layer CNN that takes the feature volume $F$ and outputs heat maps $H_j$ corresponding to the $J=21$ joints.
As in \cite{Wei2016CVPR}, a pixel in $H_j$ encodes the confidence of that pixel being covered by joint $j$.
The heat maps are used for 2D pose estimation, which is a subtask, supervised by ground truth 2D annotations.
Thus, the feature extractor and the 2D detector can be trained with 2D labeled real image data from the internet.
This drastically improves generalization ability since during training both feature extractor and 2D detector see in-the-wild images that contain more variations than images from 3D annotated datasets.
%
%
\par \noindent \textbf{3D Detector.}
Now, the 3D detector takes the feature maps $F$ and the heat maps $H$, and estimates 3D hand joint positions in the form of \textit{location maps} $L$, similar to \cite{Mehta2017ToG}.
For each joint $j$, $L_j$ has the same 2D resolution as $H_j$, and each pixel in $L_j$ encodes joint $j$'s 3D coordinates.
This redundancy helps to the robustness.
Similar to $L$, we also estimate delta maps $D$ where each pixel in $D_b$ encodes the orientation of bone $b$, represented by a 3D vector from the parent joint to the child joint.
This intermediate representation is needed to explicitly inform the network about the relation of neighboring joints in the kinematic chain.
In the 3D detector, we first use a 2-layer CNN to estimate the delta maps $D$ from the heat maps $H$ and feature maps $F$.
Next, heat maps $H$, feature maps $F$, and delta maps $D$ are concatenated and fed into another 2-layer CNN to obtain the final location maps $L$.
The location maps $L$ and the delta maps $D$ are supervised by 3D annotations.
During inference, the 3D position of joint $j$ can be retrieved by a simple look-up in the location map $L_j$ at the uv-coordinate corresponding to the maxima of the heat map $H_j$.
To alleviate the fundamental depth-scale ambiguity in the monocular setting, the predicted coordinates are relative to a root joint and normalized by the length of a reference bone.
We select the middle metacarpophalangeal to be the root joint, and the bone from this joint to the wrist is defined as the reference bone.
%
%
\par \noindent \textbf{Loss Terms.}
Our loss function
\begin{equation}
 \label{eq:L_det}
 	\mathcal{L}_\mathrm{heat} +
 	\mathcal{L}_\mathrm{loc} +
 	\mathcal{L}_\mathrm{delta} +
 	\mathcal{L}_\mathrm{reg}
\end{equation}
\noindent comprises four terms to account for the multi-task learning scheme.
%
%
First, $\mathcal{L}_\mathrm{heat}$ is defined as
\begin{equation}
 	\mathcal{L}_\mathrm{heat} = ||H^\mathrm{GT} - H||_{F}^{2}
\end{equation}
which ensures that the regressed heatmaps $H$ are close to the ground truth heatmaps $H^\mathrm{GT}$.
$||\cdot||_{F}$ denotes the Frobenius norm.
To generate the ground truth heat maps $H_j^\mathrm{GT}$ for joint $j$, we smooth $H_j^\mathrm{GT}$ with a Gaussian filter centered at the 2D annotation using a standard deviation of $\sigma = 1$.
Again note that $\mathcal{L}_\mathrm{heat}$ only requires 2D annotated image datasets.
We particularly stress the importance of such images, as they contain much more variation than those with 3D annotations.
Thus, this loss supervises our feature extractor and our 2D detector to learn the important features for hand joint detection on in-the-wild images.
%
%
To supervise the 3D detector, we propose two additional loss terms
\begin{equation}
 \label{eq:L_loc}
  \mathcal{L}_\mathrm{loc} =
  || H^\mathrm{GT} \odot (L^\mathrm{GT} - L) ||_{F}^{2}
\end{equation}
\begin{equation}
\label{eq:L_delta}
  \mathcal{L}_\mathrm{delta} =
  || H^\mathrm{GT} \odot (D^{\mathrm{GT}} - D) ||_{F}^{2}
\end{equation}
which measure the difference between ground truth and predicted location maps $L$ and delta maps $D$, respectively.
Ground truth location maps $L^\mathrm{GT}$ and delta maps $D^\mathrm{GT}$ are constructed by tiling the coordinates of the ground truth joint position and bone direction to the size of the heat maps.
Since we are mainly interested in the 3D predictions at the maxima of the heat maps, the difference is weighted with $H^\mathrm{GT}$, where $\odot$ is the element-wise matrix product.
$\mathcal{L}_\mathrm{reg}$ is a $L2$ regularizer for the weights of the network to prevent overfitting.
During training, data with 2D and 3D annotations are mixed in the same batch, and all the components are trained jointly.
Under this multi-task scheme, the network learns to predict 2D poses under diverse real world appearance from 2D labeled images, as well as 3D spatial information from 3D labeled data.
%
%
\par
\noindent \textbf{Global Translation.}
If the camera intrinsics matrix $K$ and the reference bone length $l_\mathrm{ref}$ are provided, the absolute depth $z_\mathrm{r}$ of the root joint can be computed by solving
\begin{equation}
  l_\mathrm{ref} = || K^{-1} z_\mathrm{r}\begin{bmatrix} u_\mathrm{r} \\ v_\mathrm{r} \\ 1 \end{bmatrix} - K^{-1} (z_\mathrm{r}
  + l_\mathrm{ref} * d_\mathrm{w}) \begin{bmatrix} u_\mathrm{w} \\ v_\mathrm{w} \\ 1 \end{bmatrix} ||_2
\end{equation}
Here, subscripts $\cdot_\mathrm{r}$ and $\cdot_\mathrm{w}$ denote the root and wrist joint, respectively.
$u$ and $v$ are the 2D joint predictions in the image plane and $d_\mathrm{w}$ is the normalized and root-relative depth of the wrist regressed by DetNet.
As $z_\mathrm{r}$ is the only unknown variable, one can solve for it in closed form.
After computing $z_\mathrm{r}$, the global translation in $x$ and $y$ dimension can be computed via the camera projection formula.
\subsection{Hand Model and Shape Estimation}
\label{sec:handmodel_shape}
%
%
\par \noindent \textbf{Hand Model.}
We choose MANO \cite{Romero2017ToG} as the hand model to be driven by the output of our IKNet.
The surface mesh of MANO can be fully deformed and posed by the shape parameters $\beta \in \mathbb{R}^{10}$ and pose parameters $\theta \in \mathbb{R}^{21 \times 3}$.
More specifically, $\beta$ represents the coefficients of a shape PCA bases which is learned from hand scans, while $\theta$ represents joint rotations in axis-angle representation.
They allow to deform the mean template $\bar{T} \in \mathbb{R}^{V \times 3}$ to match the shape of different identities as well as to account for pose-dependent deformations.
Here, $V$ denotes the number of vertices.
Before posing, the mean template $\bar{T}$ is deformed as
\begin{equation}
  \mathcal{T}(\beta, \theta) = \bar{T} + \mathcal{B}_{s}(\beta) + \mathcal{B}_{p}(\theta)
\end{equation}
where $\mathcal{B}_{s}(\beta)$ and $\mathcal{B}_{p}(\theta)$ are shape and pose blendshapes, respectively.
Then the posed hand model $\mathcal{M}(\theta, \beta) \in \mathbb{R}^{V \times 3}$ is defined as
\begin{equation}
\label{eq:mano}
  \mathcal{M}(\theta, \beta) = W(\mathcal{T}(\theta, \beta), \theta, \mathcal{W}, \mathcal{J}(\theta))
\end{equation}
where $W(\cdot)$ is a standard linear blend skinning function that takes the deformed template mesh $\mathcal{T}(\beta, \theta)$, pose parameters $\theta$, skinning weights $\mathcal{W}$, and posed joint locations $\mathcal{J}(\theta)$.
%
\par \noindent \textbf{Shape Estimation.}
Since we are not only interested in the pose of the hand but also its shape, we utilize the predicted joint positions to estimate shape parameters $\beta$ of the MANO model.
As the predictions are scale-normalized, the estimated shape can only represent relative hand shape, e.g., the ratio of fingers to palm.
We compute the hand shape $\beta$ by minimizing
\begin{equation}
  E(\beta) = \sum_{b}|| \frac{l_{b}(\beta)}{l_\mathrm{ref}(\beta)} - l_b^\mathrm{pred} ||_{2}^{2}+ {\lambda}_{\beta} ||\beta||_{2}^{2} \mathrm{.}
\end{equation}
Here, the first term ensures that for every bone $b$ the bone length of the deformed hand model $l_b(\beta)$ matches the length of the predicted 3D bone length $l_b^\mathrm{pred}$, that can be derived from the 3D predictions of DetNet.
Label $\cdot_\mathrm{ref}$ refers to the reference bone of the deformed MANO model.
The second term acts as a $L2$ regularizer on the shape parameters and is weighted by $\lambda_{\beta}$.
\subsection{Inverse Kinematics Network IKNet}
\label{sec:method:iknet}
Although 3D joint locations can explain the hand pose, such a representation is not sufficient to animate hand mesh models, which is important for example in computer graphics (CG) applications.
In contrast, a widely-used representation to drive CG characters are joint rotations.
We therefore infer in the network joint rotations from joint locations, also known as the inverse kinematics (IK) problem.
To this end, we propose a novel end-to-end neural network, IKNet, to solve the inverse kinematics problem.
The main benefits of our \emph{learning based} IKNet are:
First, our design allows us to incorporate MoCap data as an additional data modality to provide full supervision during training.
This is in stark contrast to methods that directly regress rotations from the image \cite{Zhang2019ICCV,Baek2019CVPR,Boukhayma2019CVPR} which only allow weakly supervised training.
Second, we can solve the IK problem at much higher speed since we only require a single feed-forward pass compared to iterative model fitting methods \cite{Mueller2018CVPR,Panteleris2018WACV}.
Third, hand pose priors can be directly learned from the data in contrast to hand-crafted priors in optimization based IK \cite{Mueller2018CVPR,Panteleris2018WACV}.
Finally, we also show that our IKNet can correct noisy 3D predictions of DetNet and the joint rotation representation is by nature bone-scale preserving.
The similar idea of an IK network was also proposed in \cite{holden2018robust}, but was used for denoising marker-based MoCap data, while we perform hand pose estimation.
%
%
\par \noindent \textbf{MoCap Data.}
When it comes to training the IKNet, one ideally wants to have paired samples of 3D hand joint positions and the corresponding joint rotation angles.
The MANO model comes with a dataset that contains 1554 poses of real human hands from 31 subjects.
Originally, the rotations are in the axis-angle representation and we convert them to the quaternion representation, which makes interpolation between two poses easier.
However, this dataset alone would still not contain enough pose variations.
Therefore, we augment the dataset based on two assumptions:
\emph{1)} we assume the pose of each finger is independent of other fingers;
\emph{2)} any interpolation in quaternion space from the rest pose to a pose from the extended dataset, that is based on 1), is valid.
Based on \emph{1)}, we choose independent poses for each finger from the original dataset and combine them to form unseen hand poses.
Based on \emph{2)}, we can now interpolate between the rest pose and the new hand poses.
To account for different hand shapes, we also enrich the dataset by sampling $\beta$ with the normal distribution $\mathcal{N}(0, 3)$.
Following the above augmentation technique, we produce paired joint location and rotation samples on-the-fly during training.
%
%
\par \noindent \textbf{3DPosData.}
However, if we train IKNet purely on this data, it is not robust with respect to the noise and errors that are contained in the 3D predictions of DetNet.
This is caused by the fact that the paired MoCap data is basically noiseless.
Therefore, we also leverage the 3D annotated image data.
In particular, we let the pre-trained DetNet produce the 3D joint predictions for all the training examples with 3D annotations and use those joint predictions as the input to the IKNet.
The estimated joint rotations of the IKNet are then passed through a forward kinematic layer to reconstruct the joint positions, which are then supervised by the corresponding ground truth 3D joint annotations.
In other words, we additionally construct a dataset with paired 3D DetNet predictions and ground truth 3D joint positions, which is used as a weak supervision to train the IKNet.
We refer to this dataset as 3DPosData in the following.
In this way, the IKNet learns to handle the 3D predictions of DetNet and is robust to noisy input.
%
%
\par \noindent \textbf{Network Design.}
We design the IKNet as a 7-layer fully-connected neural network with batch normalization, and use sigmoid as the activation function except for the last layer that uses a linear activation.
We encode the input 3D joint positions as
$\mathcal{I} = [\mathcal{X}, \mathcal{D}, \mathcal{X}_\mathrm{ref}, \mathcal{D}_\mathrm{ref}] \in \mathbb{R}^{4 \times J \times 3}$,
where $\mathcal{X}$ are the root-relative scale-normalized 3D joint positions as in Sec.~\ref{sec:method:detnet};
$\mathcal{D}$ is the orientation of each bone, which we additionally provide as input to explicitly encode information of neighboring joints.
$\mathcal{X}_\mathrm{ref}$, $\mathcal{D}_\mathrm{ref}$ encode information about the shape identity and are defined as the 3D joint positions and bone orientations in the rest pose, respectively.
They can be measured in advance for better accuracy, or inferred from the predictions of the DetNet, as described in Sec.\ref{sec:handmodel_shape}.
The output of the IKNet is the global rotation of each joint represented as a quaternion $\hat{\mathcal{Q}} \in \mathbb{R}^{J \times 4}$, which is then normalized to be a unit quaternion $\mathcal{Q}$.
We prefer the quaternion representation over an axis angle one due to the better interpolation properties that are required in our data augmentation step.
Additionally, quaternions can be converted to rotation matrices, as later used in our losses, without using trigonometric functions which are more difficult to train since they are non-injective.
To apply the final pose to the MANO model, we convert the quaternions $\mathcal{Q}$ back to the axis-angle representation, and then deform the model according to Eq.~\ref{eq:mano}.
%
%
\par \noindent \textbf{Loss Terms.}
Our loss function comprises four terms
\begin{equation}
  \label{eq:L_ik}
  	\mathcal{L}_\mathrm{cos} +
	\mathcal{L}_\mathrm{l2} +
   \mathcal{L}_\mathrm{xyz} +
	\mathcal{L}_\mathrm{norm} \mathrm{.}
\end{equation}
First, $\mathcal{L}_\mathrm{cos}$ measures the distance between the cosine value of the difference angle, which is spanned by the ground truth quaternion $\mathcal{Q}^\mathrm{GT}$ and our prediction $\mathcal{Q}$, as
\begin{equation}
  \mathcal{L}_\mathrm{cos} = 1 - real(\mathcal{Q}^\mathrm{GT} \ast \mathcal{Q}^{-1}) \mathrm{.}
\end{equation}
$real(\cdot)$ takes the real part of the quaternion, $\ast$ is the quaternion product, and $\mathcal{Q}^{-1}$ is the inverse of quaternion $\mathcal{Q}$.
Further, $\mathcal{L}_\mathrm{l2}$ directly supervises the predicted quaternion $\mathcal{Q}$:
\begin{equation}
  \mathcal{L}_\mathrm{l2} = || \mathcal{Q}^\mathrm{GT} - \mathcal{Q} ||_{2}^{2} \mathrm{.}
\end{equation}
The proposed two losses can only be applied on the MoCap data.
To also use 3DPosData, we propose a third loss, $\mathcal{L}_\mathrm{xyz}$, to measure the error in terms of 3D coordinates after posing
\begin{equation}
  \mathcal{L}_\mathrm{xyz} = || \mathcal{X}^\mathrm{GT} - FK(\mathcal{Q}) ||_{2}^{2}
\end{equation}
where $FK(\cdot)$ refers to the forward kinematics function and $\mathcal{X}^\mathrm{GT}$ is the ground truth 3D joint annotation.
Finally, to softly constrain the un-normalized output $\hat{\mathcal{Q}}$ to be unit quaternions, we apply $\mathcal{L}_\mathrm{norm}$ as
\begin{equation}
  \mathcal{L}_\mathrm{norm} = |1 - ||\hat{\mathcal{Q}}||_\mathrm{2}^{2} | \mathrm{.}
\end{equation}
\section{Results}
\label{sec:experiments}
In this section, we first provide implementation details (Sec.~\ref{sec:implementation}).
Then, we show qualitative results on challenging examples (Sec.~\ref{sec:qualitative}).
Finally, we compare our method to previous work (Sec.~\ref{sec:comparison}) and perform an ablation study to evaluate the importance of all our design choices (Sec.~\ref{sec:ablation}).
%
%
\subsection{Implementation Details}
\label{sec:implementation}
All our experiments are performed on a machine with NVIDIA GTX1080Ti graphics card, where DetNet takes 8.9$ms$ and IKNet takes 0.9$ms$ for a single feed-forward pass.
Thus, we achieve a state-of-the-art runtime performance of over 100$fps$.
%
%
\par \noindent \textbf{Training Data.}
Our DetNet is trained on 3 datasets: the CMU Panoptic Dataset (CMU) \cite{Simon2017CVPR}, the Rendered Handpose Dataset (RHD) \cite{Zimmermann2017ICCV} and the GANerated Hands Dataset (GAN) \cite{Mueller2018CVPR}.
The CMU dataset contains 16720 image samples with 2D annotations gathered from real world.
RHD and GAN are both synthetic datasets that contain 41258 and 330000 images with 3D annotations, respectively.
Note that DetNet is trained without any real images with 3D annotations.
We found that the real image 3D datasets do not contain enough variations and let our network overfit resulting in poor generalization across different datasets.
To train the IKNet, we leverage the MoCap data from the MANO model and the 3DPosData, as discussed before.
%
%
\subsection{Qualitative Results}
\label{sec:qualitative}
\begin{figure}[t]
	\centering
	\includegraphics[width=\linewidth]{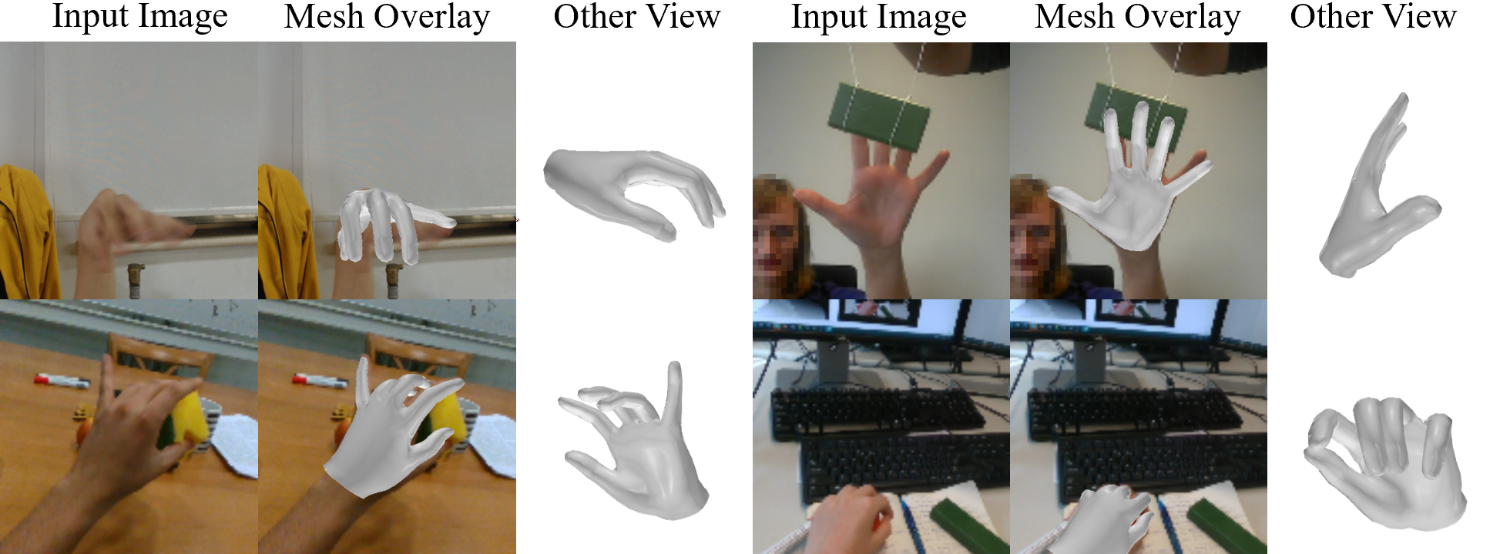}
	\caption{
		We demonstrate our results under several challenging scenarios: motion blur, object occlusion, complex pose, and unconstrained viewpoint.
		We show our results overlayed onto the input image and from a different virtual camera view.
	}
	\label{fig:qualitative}
	\vspace{-2mm}
\end{figure}

\begin{figure}[t]
	\centering
	\includegraphics[width=\linewidth]{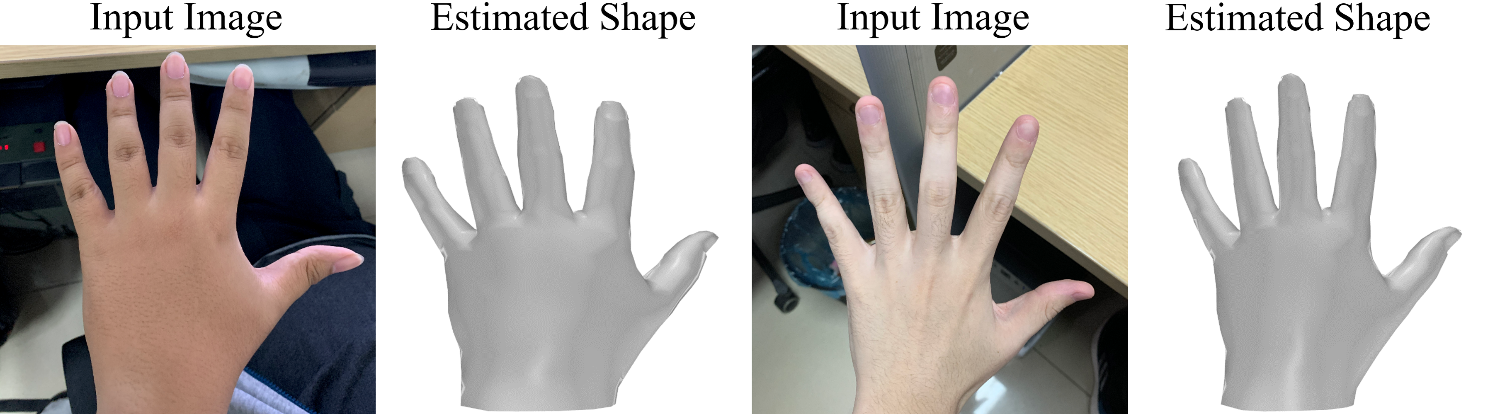}
	\caption{
		Illustration of our shape results.
		Note that our recovered shapes look visually plausible and reflect the overall shape of the subject's hand in the input image.
	}
	\label{fig:shape}
	\vspace{-2mm}
\end{figure}

\begin{figure}[t]
	\centering
	\includegraphics[width=\linewidth]{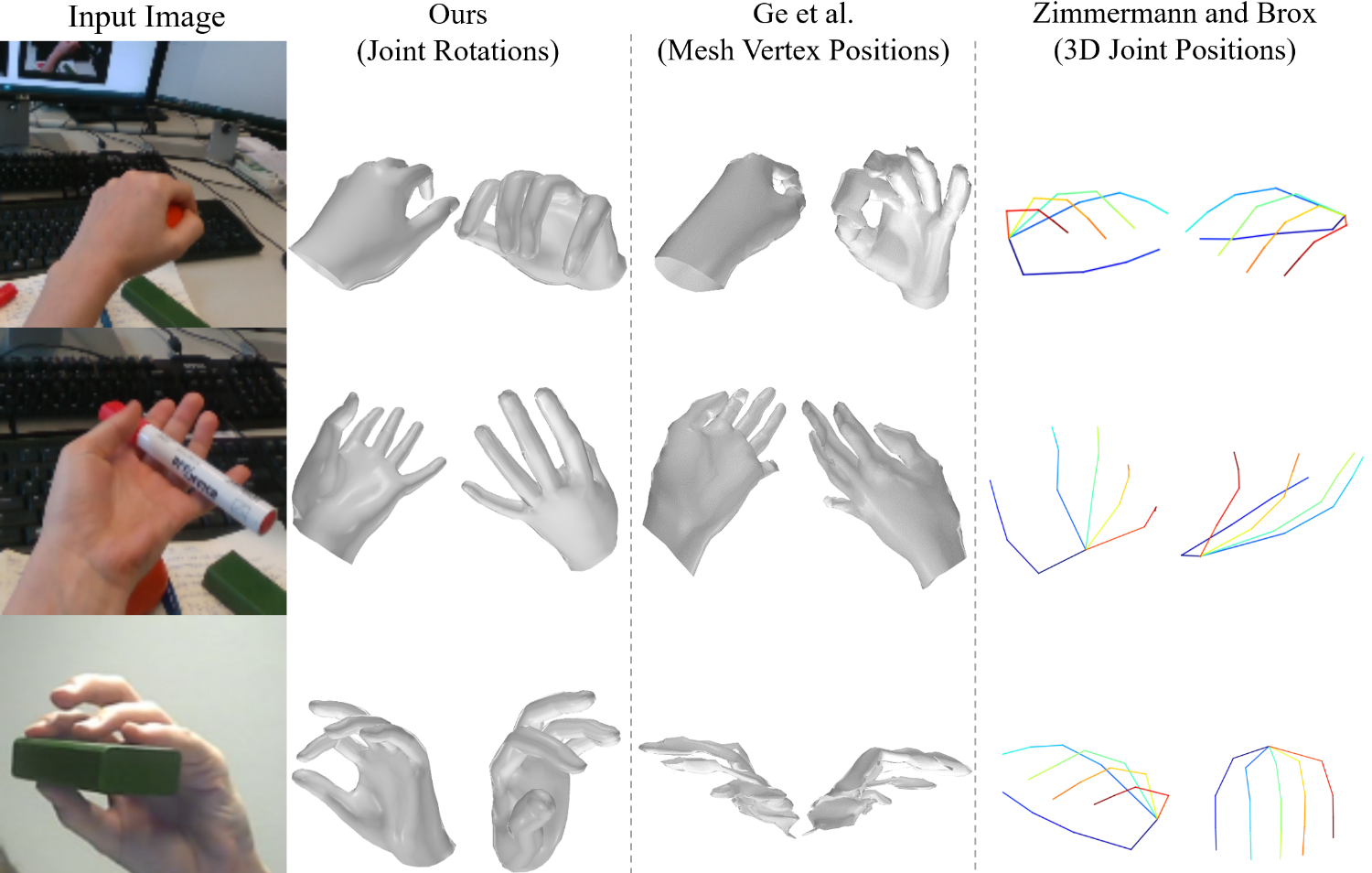}
	\caption{
		Comparison with \cite{Zimmermann2017ICCV} and Ge et al.~\cite{Ge2019CVPR}.
		Our approach cannot only output a fully deformed and posed dense 3D hand model, but also shows better robustness under occlusions compared to previous work.
		We show the same pose rendered from original and different camera view.
	}
	\label{fig:qualitative_compare}
\end{figure}

In Fig.~\ref{fig:qualitative}, we show results of our novel method on several challenging in-the-wild images demonstrating that it generalizes well to unseen data.
Most importantly, we not only predict 3D joint positions but also joint angles, allowing us to animate a hand surface model directly.
Such an output representation is much more useful in many applications in graphics and vision.
Further Fig.~\ref{fig:qualitative} demonstrates that our method works well for very fast motions and blurred images (top left), as well as complex poses such as grasping (bottom left).
Occlusions by objects (top right), self occlusions and challenging view points (bottom right) can also be handled.
More results are shown in our supplemental material.
%
%
In Fig.~\ref{fig:shape}, we demonstrate that our approach can capture different hand shapes just from a single image.
Note that finger and palm shape are correctly adjusted and they look plausible.
%
%
In Fig~\ref{fig:qualitative_compare}, we qualitatively compare our results to Zimmermann and Brox \cite{Zimmermann2017ICCV} and Ge et al.~\cite{Ge2019CVPR} on challenging images.
While \cite{Zimmermann2017ICCV} only recovers 3D joint positions, our method can animate a full 3D hand mesh model due to the joint rotation representation.
We also demonstrate superior robustness compared to ~\cite{Ge2019CVPR}, which we attribute to the combined training on 2D labeled in-the-wild images and the MoCap data.
%
%
\subsection{Comparison to Related Work}
\label{sec:comparison}
%
%
\par \noindent \textbf{Evaluation Datasets and Metrics.}
We evaluate our approach on four public datasets: the test sets of RHD \cite{Zimmermann2017ICCV} and Stereo Hand Pose Tracking Benchmark (STB) \cite{Zhang2017ICIP}, Dexter+Object (DO) \cite{Sridhar2016ECCV} and EgoDexter (ED) \cite{Mueller2017ICCV}.
Again, note that RHD is a synthetic dataset.
The STB dataset contains 12 sequences of a unique subject with 18000 frames in total.
Following \cite{Mueller2018CVPR}, we evaluate our model on 2 sequences.
The DO dataset comprises 6 sequences of 2 subjects interacting with objects from third view.
The ED dataset is composed of 4 sequences of 2 subjects performing hand-object interactions in the presence of occlusions captured from an egocentric view.
We use the following evaluation metrics: the percentage of correct 3D keypoints (PCK), and the area under the PCK curve (AUC) with thresholds ranging from 20$mm$ to 50$mm$.
As previous work, we perform a global alignment to better measure the local hand pose.
For ED and DO we aligned the centroid of the finger tip predictions to the GT one; for RHD and STB we aligned our root to the ground truth root location.
%
%
\par \noindent \textbf{Quantitative Comparison.}
%
%
In Table.~\ref{tab:stoa}, we compare our approach to other state-of-the-art methods.
Note that \emph{not} all the methods were trained on the exact same data.
Some methods use additional data, some of which not publicly available, for higher accuracy, including: synthetic images with ground truth hand mesh \cite{Ge2019CVPR}, depth images \cite{Yang2019ICCV, Cai2018ECCV,Spurr2018CVPR}, real images with 2D annotations \cite{Iqbal2018ECCV}, and real images with 3D labels from a panoptic stereo \cite{Xiang2019CVPR}.
%
%
We argue among all test datasets, the most fair comparison can be reported on the DO and ED dataset since no model used them for training.
This further means that the evaluation on DO and ED gives a good estimate of how well models generalize.
On DO and ED, our approach outperforms others by a large margin.
This is due to our novel architecture that allows combining all available data modalities, including 2D and 3D annotated image datasets as well as MoCap data.
We further stress the importance of the dataset combination used to train our model.
%
%
\par
On STB, our accuracy is within the range of our results on DO and ED, further proving that our approach generalizes across datasets.
While we achieve a worse accuracy on STB compared to others, note that our final model is \emph{not} trained on STB in contrast to all other approaches.
As many works have mentioned \cite{Xiang2019CVPR, Yang2019ICCV, Zhang2019ICCV, Iqbal2018ECCV}, the STB dataset is easily saturated.
Models tend to overfit to STB due to its large amount of frames and little variation.
We argue that the utilization of STB for training would make the training data imbalanced and harm the generalization.
This is evidenced by our additional experiment where we add STB to our training set and achieve an AUC of 0.991 on the test set of STB which is on par with previous work, but this model suffers from a huge performance drop on all other three benchmarks.
Therefore, we did not use STB to train our final model.
%
\par
For RHD, again our model achieves a consistent result as on other benchmarks.
As a synthetic dataset, RHD has different appearance and pose distribution compared to real datasets.
Previous work accounts for this by exclusively training or fine-tuning on RHD leading to superior results.
Our final model avoids this since generalization to real images is harmed.
To still proof that our architectural design is on par or better than state-of-the-art models, we made another evaluation where also exclusively train on RHD and achieve an AUC of 0.893 that is in the same ball park with others.
\begin{table}[t]
  \centering
  \resizebox{\linewidth}{!}{
    \begin{tabular}{ |c|c|c|c|c| }
    \hline
    \multirow{2}{*}{Method} & \multicolumn{4}{c|}{AUC of PCK} \\
    \cline{2-5}
                                              & DO            & ED            & STB               &  RHD          \\
    \hline
    \textbf{Ours}                                      & \textbf{.948} & \textbf{.811} & .898              & .856\textsuperscript{*}          \\
    \hline
    Ge et al.~\cite{Ge2019CVPR}               &   -           &   -           & \textbf{.998}\textsuperscript{*} & .920\textsuperscript{*}          \\
    \hline
    Zhang et al.~\cite{Zhang2019ICCV}         & .825          &   -           & .995\textsuperscript{*}          & .901\textsuperscript{*}          \\
    \hline
    Yang et al.~\cite{Yang2019ICCV}           &   -           &   -           & .996\textsuperscript{*}          & \textbf{.943}\textsuperscript{*} \\
    \hline
    Baek et al.~\cite{Baek2019CVPR}           & .650          &   -           & .995\textsuperscript{*}          & .926\textsuperscript{*}          \\
    \hline
    Xiang et al.~\cite{Xiang2019CVPR}         & .912          &   -           & .994\textsuperscript{*}          &   -           \\
    \hline
    Boukhayma et al.~\cite{Boukhayma2019CVPR} & .763          & .674          & .994\textsuperscript{*}          &   -           \\
    \hline
    Iqbal et al.~\cite{Iqbal2018ECCV}         & .672          & .543          & .994\textsuperscript{*}          &   -           \\
    \hline
    Cai et al.~\cite{Cai2018ECCV}             &   -           &   -           & .994\textsuperscript{*}          & .887\textsuperscript{*}          \\
    \hline
    Spurr et al.~\cite{Spurr2018CVPR}         & .511          &   -           & .986\textsuperscript{*}          & .849\textsuperscript{*}          \\
    \hline
    Mueller et al.~\cite{Mueller2018CVPR}     & .482          &   -           & .965\textsuperscript{*}          &   -           \\
    \hline
    Z\&B \cite{Zimmermann2017ICCV}            & .573          &   -           & .948\textsuperscript{*}          & .670\textsuperscript{*}          \\
    \hline
    \end{tabular}
  }
  \caption{
    Comparison with state-of-the-art methods on four public datasets.
    We use "*" to note that the model was trained on the dataset, and use "-" for those who did not report the results.
    Our system outperforms others by a large margin on the DO and ED dataset which we argue is the most fair comparison as none of the models are trained on these datasets.
    As \cite{Iqbal2018ECCV} only reports results without alignment, we report the absolute values for this method.
  }
  \label{tab:stoa}
\end{table}
%
%
\subsection{Ablation Study}
\label{sec:ablation}
\begin{figure}[t]
	\centering
	\includegraphics[width=\linewidth]{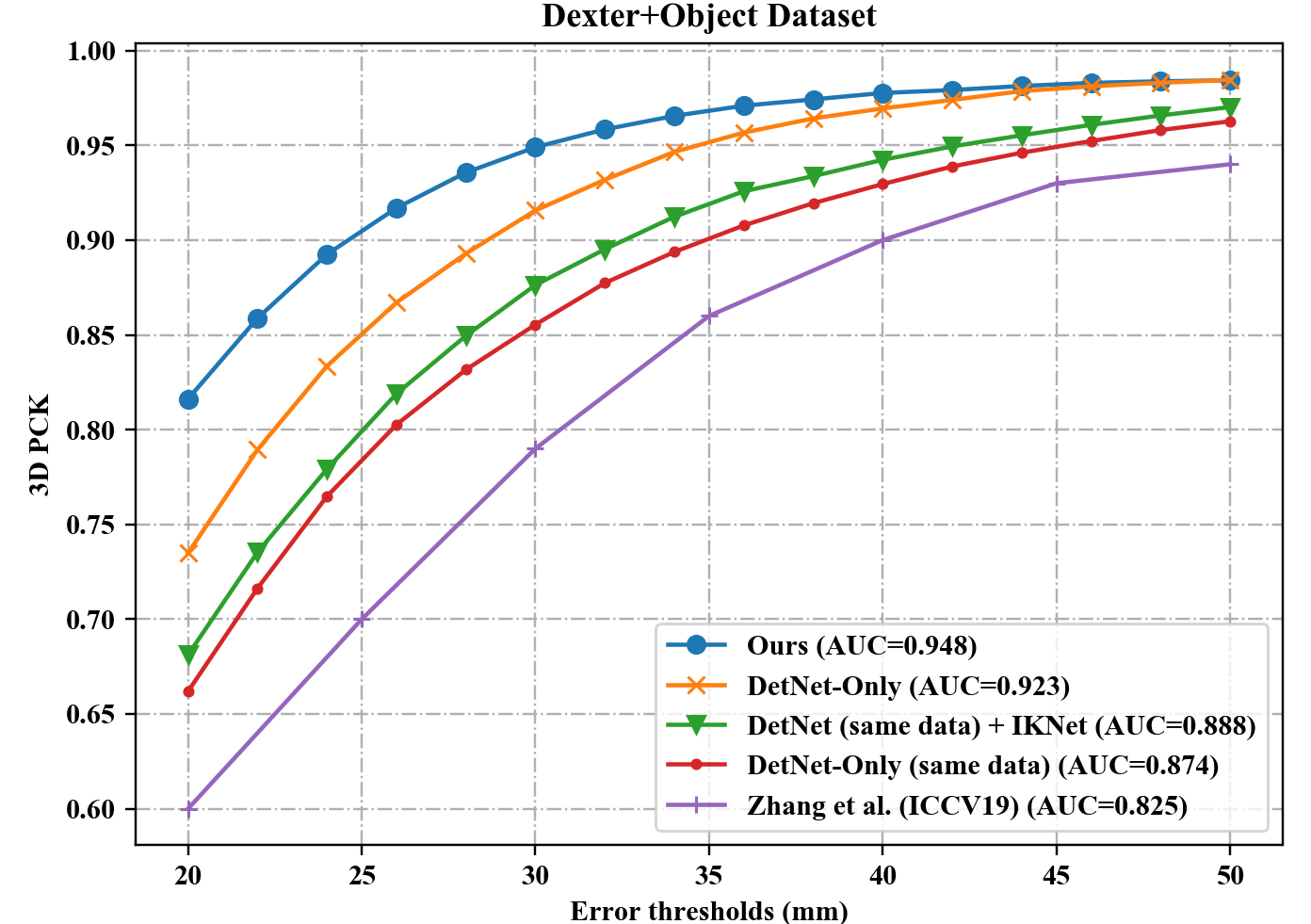}
	\caption{
		Ablation study for the training data on DO.
		We use "same data" to indicate our model trained with RHD and STB, which is the same as Zhang et al.~\cite{Zhang2019ICCV}.
		We demonstrate that our architecture is superior to theirs by design.
		Integrating more data further boosts the results.
	}
	\label{fig:do_data_abl}
	\vspace{-2mm}
\end{figure}

\begin{figure}[t]
	\centering
	\includegraphics[width=\linewidth]{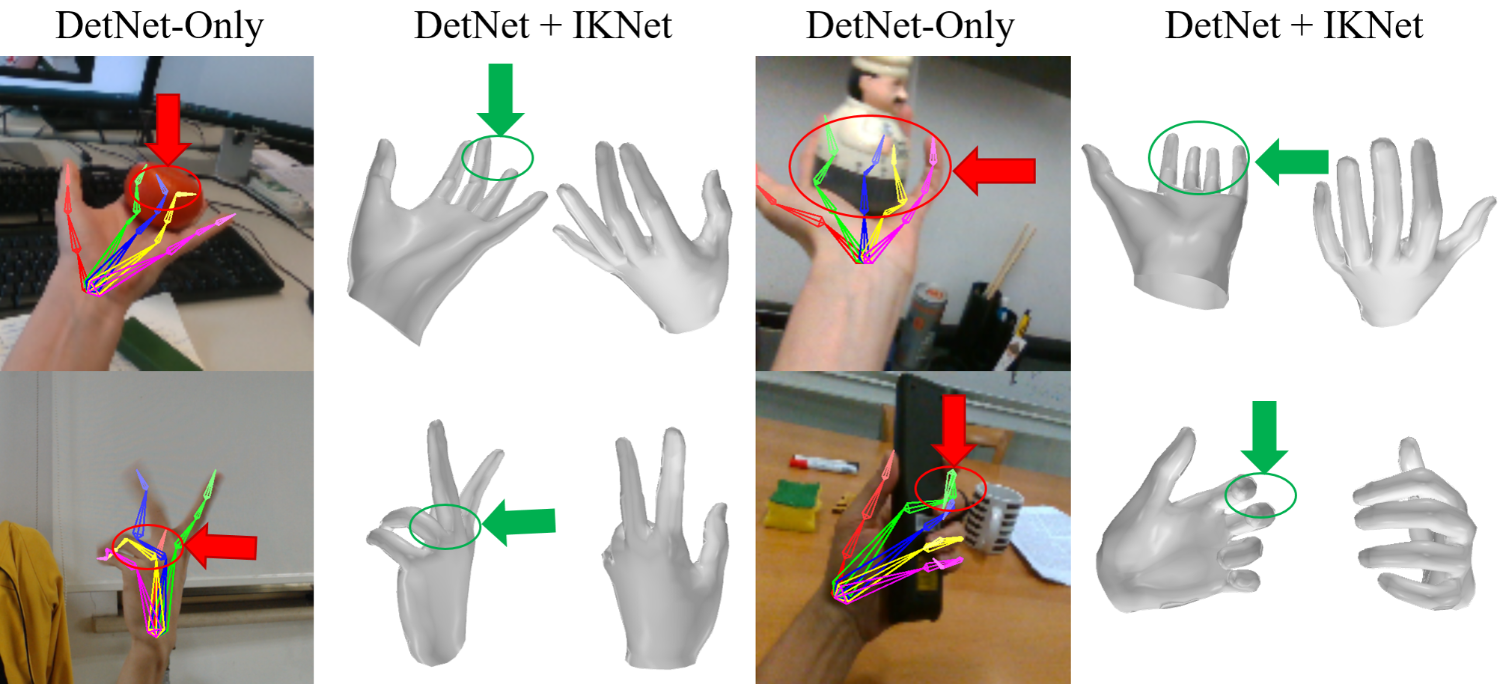}
	\caption{
		Our IKNet is able to compensate some errors from the DetNet based on the prior learned from the MoCap data.
	}
	\label{fig:refine}
	\vspace{-2mm}
\end{figure}

\begin{figure}[t]
	\centering
	\includegraphics[width=\linewidth]{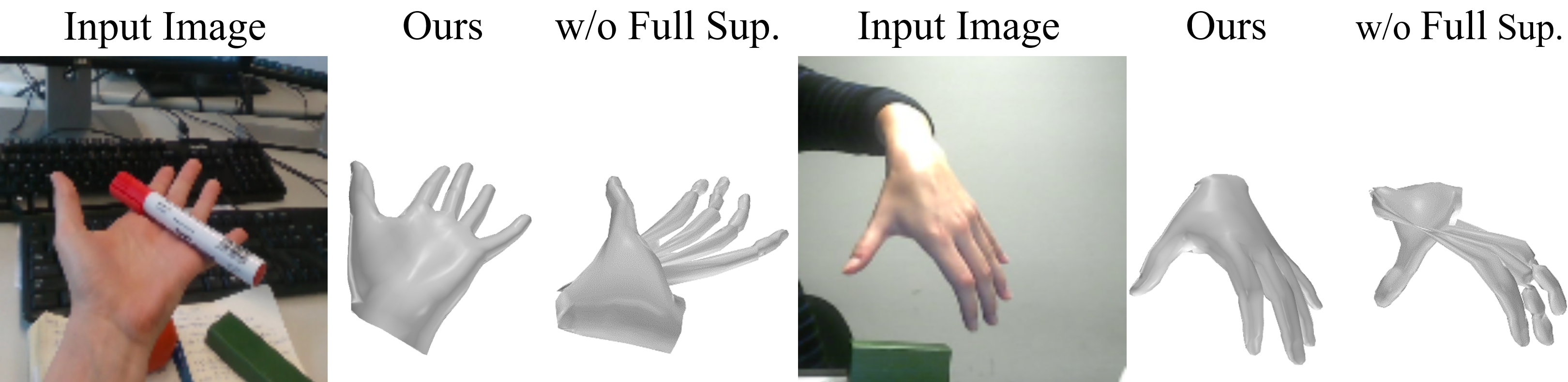}
	\caption{
		Comparison between IKNets with and without rotational supervision from MoCap data.
		Note that even though 3D joint positions match the ground truth, without this supervision unnatural poses are estimated.
	}
	\label{fig:ik_vis_wf}
	\vspace{-4mm}
\end{figure}

In Table.~\ref{tab:ablation} and Fig.~\ref{fig:do_data_abl}, we evaluate the key components of our approach:
specifically, we evaluate
\emph{1)} our architectural design and the combination of training data compared to a baseline,
\emph{2)} the impact of the IKNet over a pure 3D joint position regression of DetNet, which we refer to as DetNet-only,
\emph{3)} the influence of direct rotational supervision on joint rotations enabled by the MoCap data,
\emph{4)} how our weak supervision, using the 3DPosData, helps the IKNet to adapt to noisy 3D joint predictions,
and \emph{5)} the influence of the two loss terms on the quaternions.
%
%
\begin{table}[t]
	\centering
	\begin{tabular}{ |c|c|c|c|c| }
		\hline
		\multirow{2}{*}{} & \multirow{2}{*}{Variants of our Method} & \multicolumn{3}{c|}{AUC of PCK} \\
		\cline{3-5}
		&               & DO    & ED    & STB  \\
		\hline
		1) & \textbf{Ours}  & \textbf{.948}  & .811  & \textbf{.898} \\
		\hline
		2) & w/o IKNet                     		& .923  & .804  & .891 \\
		\hline
		3) & w/o $L_\mathrm{l2}$ and $L_\mathrm{cos}$   & .933  & \textbf{.823}  & .869 \\
		\hline
		4) & w/o 3DPosData          & .926  & .809  & .873 \\
		\hline
		5) & w/o $L_\mathrm{l2}$          	 & .943  & .812  & .890 \\
		\hline
		5) & w/o $L_\mathrm{cos}$          & .840  & .782  & .808 \\
		\hline
	\end{tabular}
	\caption{
	    Ablation study.
		We evaluate the influence of:
		\emph{2)} IKNet
		\emph{3)} Direct rotational supervision on joint rotations.
		\emph{4)} Weak supervision on joint rotations.
		\emph{5)} Loss terms on the quaternions.
	}
	\label{tab:ablation}
\end{table}
%
%
\emph{1)} As a baseline, we compare to Zhang et al.~\cite{Zhang2019ICCV} as they report state-of-the-art results on DO without using any datasets that are not publicly available (in contrast to Xiang et al.~\cite{Xiang2019CVPR} who leverage 3D annotations on CMU that are not released).
To evaluate our model architecture, we trained DetNet on exactly the same data as \cite{Zhang2019ICCV} which brings an improvement of around 5\% compared to \cite{Zhang2019ICCV}.
This shows that our architecture itself helps to improve accuracy.
Adding the IKNet, additionally trained on MoCap data, further improves the result.
This further proves that disentangling the motion capture task into joint detection and rotation recovery makes the model easier to train, and also enables to leverage of MoCap data.
Finally, the results are significantly improved with the proposed combination of training data, especially the in-the-wild 2D-labeled images.
%
%
%
%
%
\emph{2)} Across all datasets, IKNet improves the over DetNet-only.
This can be explained as our IKNet acts like a pose prior, learned from MoCap data, and can therefore correct raw 3D joint predictions of DetNet.
In Fig.~\ref{fig:refine}, the DetNet itself cannot estimate the 3D joint positions correctly.
Nevertheless, our learned hand pose prior, built-in the IKNet, can correct those wrong predictions.
%
%
%
%
%
\emph{3)} Here, we removed all rotational supervision terms and only use weak supervision.
Despite the numerical results are on par with our final approach, the estimated rotations are anatomically wrong, as shown in Fig~\ref{fig:ik_vis_wf}.
This indicates that adding rotational supervision, retrieved from the MoCap data, makes training much easier and leads to anatomically more correct results.
%
%
%
%
%
\emph{4)} The difference between 1) and 4) in Table.~\ref{tab:ablation} demonstrates that the 3DPosData is crucial to make the IKNet compatible to the DetNet.
In order words, without this data the IKNet never sees the noisy 3D predictions of the DetNet but only the accurate 3D MoCap data.
Thus, it even makes the results worse.
Feeding the IKNet with the output of the pre-trained DetNet helps to deal with the noisy 3D predictions and achieves the best results.
%
%
\emph{5)} Finally in terms of network training, we found that $L_\mathrm{cos}$ is a better metric to measure the difference between two quaternions compared to the naive $L_\mathrm{l2}$ and the combination of the two gives the highest accuracy on average.

\section{Conclusion}
We proposed the first learning based approach for monocular hand pose and shape estimation that utilizes data from two completely different modalities: image data and MoCap data.
Our new neural network architecture features a trained inverse kinematics network that directly regresses joint rotations.
These two aspects leads to a significant improvement over state of the art in terms of accuracy, robustness and runtime.
%
%
In the future, we plan to extend our model to capture the hand texture by in-cooperating dense 3D scans.
Another direction is the joint capturing of two interacting hands from a single RGB image which currently is only possible with depth sensors.
%
%
%
{\small
	\bibliographystyle{ieee_fullname}
	\bibliography{bibtex}
}
%
%
\end{document}